\ifdefined\pdfminorversion\pdfminorversion=4\fi
\ifdefined\pdfobjcompresslevel\pdfobjcompresslevel=0\fi

\IfFileExists{ieeeconf.cls}{%
  \documentclass[a4paper,10pt,conference]{ieeeconf}%
}{%
  \documentclass[a4paper,conference]{IEEEtran}
}
\IEEEoverridecommandlockouts
\providecommand{\overrideIEEEmargins}{}
\usepackage[utf8]{inputenc}
\usepackage[T1]{fontenc}
\usepackage{amsmath,amssymb}
\usepackage{graphicx}
\usepackage{booktabs}
\usepackage{cite}
\usepackage{url}
\usepackage{xspace}
\usepackage{balance}
\usepackage{dblfloatfix}
\usepackage[caption=false,font=footnotesize]{subfig}

\newcommand{\method}{RB-TRG\xspace}

\newcommand{\includeorplaceholder}[2][]{%
  \IfFileExists{#2}{\includegraphics[#1]{#2}}{%
    \fbox{\parbox[c][32mm][c]{0.92\linewidth}{\centering
      Figure file not included in the supplied draft\\[1mm]
      \texttt{\detokenize{#2}}}}}}

\title{\LARGE \bf
Robot-Body-Aware Traversal Risk Graph Planning for\\
Wheeled-Legged Robots in Complex Terrain}

\author{Zhiqiao Guo$^{*}$, Bichi Zhang$^{*}$, and S\"oren Schwertfeger$^{\dagger}$%
\thanks{The authors are with the Key Laboratory of Intelligent Perception and Human-Machine Collaboration - ShanghaiTech University, Ministry of Education, China.  \{guozhq2022, zhangbch2025, soerensch\}@shanghaitech.edu.cn.}%
\thanks{$^{*}$Equal contribution. $^{\dagger}$Corresponding author.}%
\thanks{This work was supported by the National Natural Science Foundation of China under Grant W2531052 and the Key Laboratory of Intelligent Perception and Human-Machine Collaboration (ShanghaiTech University), Ministry of Education, China. The experiments were supported by the Core Facility Platform of Computer Science and Communication, SIST, ShanghaiTech University.}%
}

\begin{document}

\maketitle

\thispagestyle{empty}
\pagestyle{empty}

\begin{abstract}
Traversal Risk Graphs (TRGs) provide a compact, terrain-aware representation for global navigation, but native TRG costs are computed over circular node neighborhoods and edge-aligned terrain regions rather than the robot's oriented body footprint. For wheeled-legged robots, this abstraction can miss partial support loss and body--terrain interference, especially during turns. We present Robot-Body-Aware TRG planning (\method), which builds on the sparse TRG representation and lifts edge-wise terrain-risk search to heading- and turn-aware body-risk transitions. An oriented rectangular footprint is sampled along graph edges and yaw sweeps to measure longitudinal support variation, lateral inclination, terrain interference, and exposure to untrusted map regions. Mean-and-upper-tail features are incorporated into transition costs, whose accumulated value is minimized by A* over ordered node-pair states, preserving TRG construction and its planning interface. We evaluate \method in a same-graph study on four scanned terrain environments and in paired closed-loop MuJoCo trials. \method reduces the three core geometric body-placement metrics and increases end-to-end success from $51.5\%$ to $68.5\%$, while increasing mean path length by $2.3\%$. A Go2-W deployment further demonstrates \method with a full LiDAR navigation stack, which received the Best Autonomy and Best Mobility awards at the IEEE ICRA 2026 Legged Robot Challenges. The code for \method is released at \url{https://github.com/ZhiqiaoGuo/RB-TRG}.
\end{abstract}

\begin{figure}[!t]
    \centering
    \includeorplaceholder[width=0.99\columnwidth]{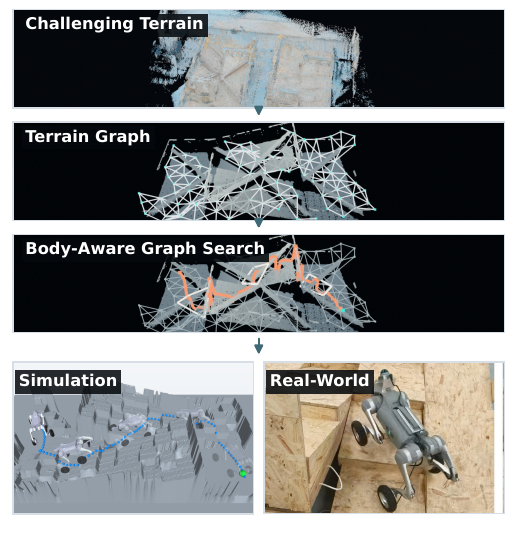}
    \caption{\method at a glance. A scanned terrain is represented by a sparse terrain graph, on which body-aware graph search selects an execution path. The resulting motion is evaluated in simulation and on the physical Go2-W platform.}
    \label{fig:overview}
\end{figure}

\section{Introduction}
\label{sec:introduction}

Wheeled-legged robots combine efficient rolling with the ability to negotiate slopes, steps, and irregular terrain using articulated legs~\cite{bjelonic2022survey,jelavic2021combined}. Their global routes must therefore satisfy more than centerline collision avoidance. A path may traverse terrain that appears safe at its center while leaving part of the robot unsupported, inducing excessive body inclination, or sweeping the chassis across an obstacle during a turn. These failure modes depend jointly on terrain geometry, robot dimensions, and heading.

Elevation maps provide an efficient representation of non-planar terrain~\cite{fankhauser2018probabilistic,miki2022elevation}. Existing navigation methods derive scalar traversability costs, stochastic risk, pose feasibility, or learned motion cost from such maps~\cite{fan2021step,wermelinger2016navigation,yang2021learned,wellhausen2023artplanner}. At the other end of the modeling spectrum, contact, configuration-space, and whole-body planners reason explicitly about robot pose, contact sequence, and system constraints~\cite{hauser2008motion,vernaza2009search,jelavic2021combined}. The former support efficient global planning but may omit finite-body placement effects; the latter provide richer feasibility reasoning at substantially higher modeling and computational cost.

TRG-Planner~\cite{lee2025trg} occupies an intermediate point in this spectrum. It constructs a sparse graph whose connectivity and edge weights encode terrain reachability and traversal risk. However, its circular node regions and edge-aligned risk model do not explicitly represent an anisotropic body along a transition or through the yaw sweep between consecutive edges. The native cost can therefore rank routes similarly even when their body-placement implications differ.

We address this gap with \method, which couples TRG's sparse terrain representation with orientation-conditioned body reasoning. A robot-sized oriented footprint is evaluated along both edges and turns. Four geometric signals---front--rear support variation, lateral inclination, internal terrain interference, and exposure to untrusted map regions---are aggregated through mean and upper-tail terms and integrated with native terrain risk. Ordered-pair search conditions each transition on the incoming heading and associated turn. The same-graph evaluation holds graph topology, native edge risk, and snapped endpoints fixed; the paired MuJoCo protocol additionally controls smoothing, locomotion, terrain, and perturbation seeds. Together, these protocols directly attribute route and execution differences to body-aware transition scoring.

The contributions of this paper are:
\begin{itemize}
    \item We formulate heading- and turn-aware body risk by evaluating the complete oriented footprint along graph edges and yaw sweeps, with explicit treatment of support variation, body inclination, terrain interference, and untrusted-map exposure.
    \item We introduce an ordered-pair TRG search that integrates mean-and-tail body risk with native terrain risk in heading-conditioned transition costs.
    \item We validate \method through controlled same-graph experiments, component ablations, paired closed-loop simulations, comparisons with reproduced baselines, and Go2-W deployment. \method changes the node sequence in 157 of 200 same-graph cases and raises end-to-end success over Native TRG by 17.0 points.
\end{itemize}

\section{Related Work}
\label{sec:related_work}

\subsection{Terrain-Aware Global Planning}

Terrain-aware global planning requires both an efficient terrain representation and a meaningful risk model. Elevation maps and related 2.5-D representations preserve local surface geometry while remaining inexpensive to query~\cite{fankhauser2018probabilistic,miki2022elevation,borges2022survey}. Risk can then be derived from geometric terrain properties, uncertainty, or learned motion cost. STEP propagates map uncertainty into stochastic risk-sensitive planning~\cite{fan2021step}; learned robot-specific motion costs support real-time route selection~\cite{yang2021learned}; and Kr{\"u}si \emph{et al.} combine terrain assessment, motion planning, and trajectory optimization directly on point clouds~\cite{krusi2017pointcloud}.

Global planning systems differ in how they organize this information. T-Hybrid combines planar obstacle information with pose-aware terrain assessment in a hybrid-map search~\cite{liu2023hybrid}. For wheeled-legged platforms, behavior-tree systems add online replanning and recovery~\cite{deluca2023navigation}, while learned navigation and locomotion policies support large-scale autonomy~\cite{lee2024robust}. Beyond geometric costs, osmAG provides hierarchical topometric structure~\cite{feng2023osmag}, and recent work uses large language models to interpret its topology and hierarchy~\cite{xie2024empowering}. TRG-Planner constructs a sparse wavefront graph with direction-dependent traversal risk~\cite{lee2025trg}. These global abstractions are effective, but their route costs do not necessarily score the complete oriented body footprint during both translation and turning.

\subsection{Robot-Scale and Morphology-Aware Planning}

Robot-scale terrain evaluation extends pointwise traversability by querying the geometry covered by a candidate pose. Wermelinger \emph{et al.} plan paths using robot-specific obstacle-negotiation capabilities~\cite{wermelinger2016navigation}, while Hines \emph{et al.} combine terrain geometry, visibility, and vehicle attitude to handle negative obstacles~\cite{hines2021virtual}. For tracked robots on difficult 3-D terrain, Yuan \emph{et al.} plan flipper configurations along a traversal rather than selecting the path itself~\cite{yuan2020configuration}.

Higher-fidelity approaches reason explicitly about reachability, contacts, or motion. Reachability-based planning captures torso and limb feasibility over rough terrain~\cite{wellhausen2021rough}, and ArtPlanner augments this abstraction with learned foothold and motion costs~\cite{wellhausen2023artplanner}. For wheeled-legged robots, Medeiros \emph{et al.} optimize base and wheel states together with contact forces~\cite{medeiros2020trajectory}. Other formulations plan footholds, body motion, contact sequences, or whole-body trajectories~\cite{hauser2008motion,vernaza2009search,jelavic2021combined,jenelten2022tamols}.

\method brings full-footprint, orientation-conditioned transition reasoning to sparse terrain graphs. It scores both translation and reorientation at graph resolution, enabling body-scale geometry to directly inform global route selection through controller-independent costs.

\section{Robot-Body-Aware TRG Planning}
\label{sec:method}

\subsection{Body-Aware Transition Formulation}

\begin{figure*}[!t]
    \centering
    \includeorplaceholder[width=0.99\textwidth]{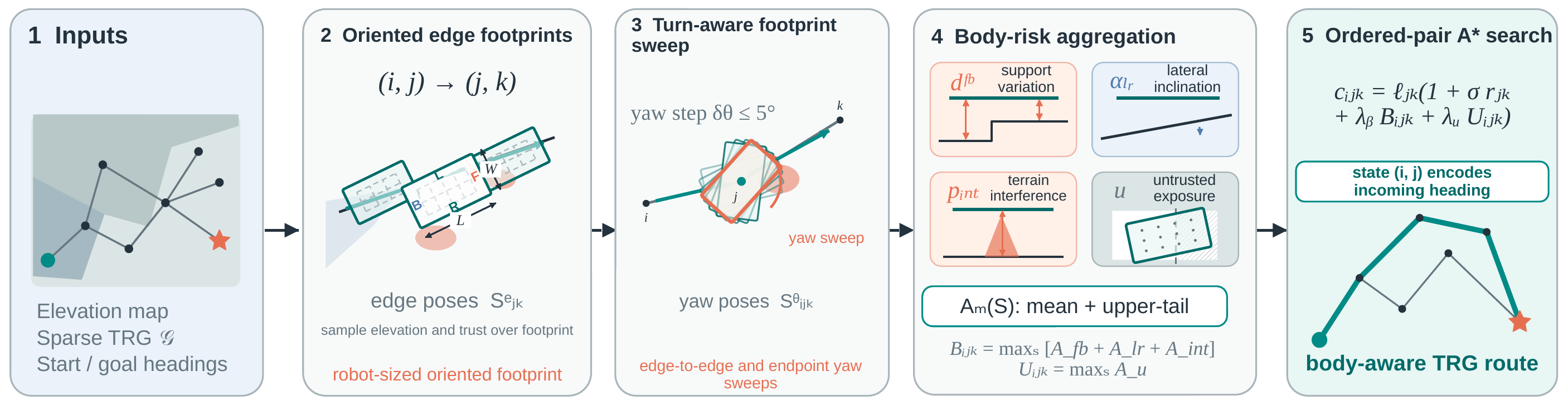}
    \caption{\method pipeline. Oriented edge and yaw-sweep footprints define mean-and-upper-tail risks for ordered-pair A* search.}
    \label{fig:method_pipeline}
\end{figure*}

Let $\mathcal{G}=(\mathcal{V},\mathcal{E})$ denote a TRG constructed as in~\cite{lee2025trg}. Each directed edge $e_{ij}\in\mathcal{E}$ has planar length $\ell_{ij}$ and native traversal risk $r_{ij}$, and the native pipeline snaps start and goal requests to valid nodes. \method builds on this sparse topology and native risk model, then recasts TRG route search over ordered node-pair states so that every transition is conditioned on its incoming heading. State $(i,j)$ represents arrival at $v_j$ from $v_i$. A transition from $(i,j)$ to $(j,k)$ evaluates both edge $e_{jk}$ and the yaw sweep from $e_{ij}$ to $e_{jk}$; initial and terminal sweeps incorporate the requested start and goal headings. Figure~\ref{fig:method_pipeline} summarizes the pipeline.

\subsection{Footprint Sampling Along Edges and Turns}

At pose $q=(\mathbf{p},\theta)$, the robot body is approximated by an oriented rectangle of length $L$ and width $W$:
\begin{equation}
\begin{aligned}
\mathcal{F}_0 &= [-L/2,L/2]\times[-W/2,W/2],\\
\mathcal{F}(q) &= \{\mathbf{p}+\mathbf{R}(\theta)\mathbf{s}\mid \mathbf{s}\in\mathcal{F}_0\}.
\end{aligned}
\label{eq:footprint}
\end{equation}
The footprint is sampled on a grid whose spacing does not exceed the elevation-map resolution. Poses along an edge are separated by at most $\delta_s$, with heading aligned to the edge direction. At each graph vertex, additional poses sample the shortest yaw rotation from the incoming to the outgoing heading at increments no larger than $\delta_\theta$, thereby capturing body placement during turning rather than only the incident edge directions.

Near map boundaries, the elevation grid is padded by the footprint circumradius. Padded cells receive finite interpolation heights but remain untrusted, contribute to the untrusted-exposure term, and are excluded from observed support.

\subsection{Body-Risk Features}

Let $H(\mathbf{f}_a)$ denote the bilinearly interpolated elevation at footprint sample $\mathbf{f}_a$, and let $M(\mathbf{f}_a)\in\{0,1\}$ indicate whether the corresponding map cell is trusted. Four corner-band means define $\bar z_F=(\bar z_{FL}+\bar z_{FR})/2$ and $\bar z_B=(\bar z_{BL}+\bar z_{BR})/2$; $\bar z_L$ and $\bar z_R$ are defined analogously. The pose-level features are
\begin{equation}
\begin{aligned}
d_{\mathrm{fb}}(q) &= |\bar z_F-\bar z_B|,\\
\alpha_{\mathrm{lr}}(q) &= \frac{180}{\pi}\,
\operatorname{atan2}(|\bar z_L-\bar z_R|,W),\\
p_{\mathrm{int}}(q) &= \max(0,z_{\max}-z_c),\\
u(q) &= 1-\frac{1}{N}\sum_{a=1}^{N}M(\mathbf{f}_a),
\end{aligned}
\label{eq:pose_metrics}
\end{equation}
where $z_{\max}$ is the maximum sampled elevation within the footprint and $z_c=\max(\bar z_{FL},\bar z_{FR},\bar z_{BL},\bar z_{BR})$. Here $d_{\mathrm{fb}}$ and $\alpha_{\mathrm{lr}}$ capture longitudinal support variation and lateral inclination, $p_{\mathrm{int}}$ measures terrain rising above the highest corner support, and $u$ is the fraction of untrusted footprint samples.

Let $g_{\mathrm{fb}}=d_{\mathrm{fb}}$, $g_{\mathrm{lr}}=\alpha_{\mathrm{lr}}$, and $g_{\mathrm{int}}=p_{\mathrm{int}}$. For $m\in\{\mathrm{fb},\mathrm{lr},\mathrm{int}\}$, the normalized feature is $x_m(q)=\min(3,g_m(q)/s_m)$; for the untrusted channel, $x_u(q)=u(q)$. Given a pose set $\mathcal{S}$ sampled along an edge or yaw sweep, feature $m$ is aggregated as
\begin{equation}
A_m(\mathcal{S}) =
 w_{\mu}\frac{1}{K}\sum_{q\in\mathcal{S}}x_m(q)
 +w_{t}\frac{1}{k}\sum_{q\in\mathcal{T}_{m,k}(\mathcal{S})}x_m(q),
\label{eq:risk_aggregate}
\end{equation}
where $K=|\mathcal{S}|$, $k=\max(1,\lceil0.1K\rceil)$, and $\mathcal{T}_{m,k}(\mathcal{S})$ contains the $k$ largest samples. The mean term penalizes persistent exposure, whereas the upper-tail term preserves short but severe events that would be diluted by averaging alone.

\subsection{Transition Cost and Search}

Let $\mathcal{S}^{e}_{jk}$ be the pose set sampled along edge $e_{jk}$, and let $\mathcal{S}^{\theta}_{ijk}$ be the associated yaw sweep; the terminal sweep is included when $k$ is the goal. The body-risk and untrusted-exposure scores for transition $(i,j)\rightarrow(j,k)$ are
\begin{equation}
\begin{aligned}
B_{ijk} &= \max_{\mathcal{S}\in\{\mathcal{S}^{e}_{jk},\mathcal{S}^{\theta}_{ijk}\}}
\left[A_{\mathrm{fb}}(\mathcal{S})+A_{\mathrm{lr}}(\mathcal{S})+A_{\mathrm{int}}(\mathcal{S})\right],\\
U_{ijk} &= \max_{\mathcal{S}\in\{\mathcal{S}^{e}_{jk},\mathcal{S}^{\theta}_{ijk}\}} A_u(\mathcal{S}).
\end{aligned}
\label{eq:transition_risk}
\end{equation}
The resulting transition cost is
\begin{equation}
c_{ijk}=\ell_{jk}\left(1+\sigma r_{jk}+\lambda_b B_{ijk}+\lambda_u U_{ijk}\right).
\label{eq:transition_cost}
\end{equation}
A deterministic A* search over ordered node pairs minimizes the accumulated cost using planar Euclidean distance to the goal as the heuristic; edge and turn features are cached within each query.

\section{Experiments}
\label{sec:experiments}

The evaluation asks four questions: (i) does body-aware scoring change route selection and geometric risk on an identical graph; (ii) what do yaw-sweep modeling and upper-tail aggregation contribute; (iii) does the full formulation improve closed-loop traversal over Native TRG; and (iv) how does \method compare with reproduced T-Hybrid and ART baselines?

\begin{table}[!t]
\caption{Same-Graph Planning Ablation}
\label{tab:static_results}
\centering
\scriptsize
\setlength{\tabcolsep}{1.4pt}
\begin{tabular}{lccccc}
\toprule
Method & $L$ [m] & $d_{\mathrm{fb}}^{\max}$ [m] &
$\alpha_{\mathrm{lr}}^{\max}$ [$^\circ$] & $p_{\mathrm{int}}^{\max}$ [m] & Untr. [\%] \\
\midrule
Native TRG & \textbf{5.059} & 0.600 & 36.4 & 0.564 & 10.4\\
Edge-only & 5.156 & 0.555 & 33.8 & 0.491 & \textbf{10.3} \\
Mean-only & 5.163 & 0.551 & 33.7 & 0.507 & 10.4 \\
Full \method & 5.173 & \textbf{0.540} & \textbf{33.3} & \textbf{0.480} & \textbf{10.3} \\
\bottomrule
\end{tabular}
\end{table}

\subsection{Terrain Data and Evaluation Protocol}

We use four RoboCup Rescue terrain scans with slopes, steps, elevated surfaces, and narrow transitions. Each point cloud is converted to a $0.02$~m elevation map. The body model uses $L=0.70$~m, $W=0.43$~m, $\delta_s=0.05$~m, and $\delta_\theta=5^\circ$. The normalization scales are $s_{\mathrm{fb}}=0.30$~m, $s_{\mathrm{lr}}=30^\circ$, and $s_{\mathrm{int}}=0.30$~m; the weights are $w_\mu=w_t=0.5$, $\lambda_b=\lambda_u=1$, and $\sigma=3$.

\textbf{Static same-graph study:} The study uses a common-valid set: for each scene, five start--goal queries are evaluated on ten shared graph seeds, yielding 200 paired instances. Endpoints are snapped only if a valid node lies within $0.50$~m. All variants use the same serialized graph, native risks, and snapped endpoints; routes are sampled at $0.02$~m and evaluated by path length, maximum front--rear variation, lateral inclination, interference, and untrusted exposure, averaged per query, per scene, and then across scenes.

The static ablation compares Native TRG; Edge-only, which uses edge poses and mean-and-upper-tail aggregation but omits vertex yaw sweeps; Mean-only, which uses edge and yaw-sweep poses but omits the upper-tail term; and Full \method, which uses all poses with mean-and-upper-tail aggregation.

\textbf{Closed-loop simulation:} We evaluate ten requested start--goal queries per scene under five friction-only trials, giving 40 queries and 200 nominal trials per method. Native TRG and \method share the endpoint-preserving cubic B-spline procedure, a $0.02$~m route sampling interval, a $0.08$~m corridor-deviation limit, and identical trial-wise friction draws. We reproduce T-Hybrid~\cite{liu2023hybrid} and the geometric ART reachability planner~\cite{wellhausen2021rough} on the same endpoints and terrains. ART searches collision-free SE(3) torso poses subject to limb-reachability volumes using the released \texttt{lazy\_prm\_star\_min\_update} configuration without learned motion cost. T-Hybrid reuses the Native--RB friction realizations, while ART uses the same friction range. All four methods use the same Unitree Go2-W model, \texttt{rl\_sar} policy at 50~Hz, $0.50$~m/s target speed, initialization checks, timeout, and success definition. Native TRG and \method constitute the paired comparison; all four methods contribute to the common task-level benchmark.

T-Hybrid applies one hard slope threshold to endpoint validity and every search primitive. We use $0.42$~rad consistently for endpoints, motion primitives, and the goal connector because the released $0.30$~rad setting rejects the requested Scene~2 endpoint.

Initialization uses route-tangent yaw, support-plane alignment, a deterministic vertical-clearance search, and 3~s of settling. Trials are execution-eligible if at least three wheels are in contact, no non-wheel contact is present, and initial roll/pitch meet the scene limits ($20^\circ$ in Scenes~1--3; $35^\circ$ roll and $45^\circ$ pitch in Scene~4). MuJoCo heightfields use the planning elevation arrays with a 1~m zero-height apron. Success requires reaching within $0.25$~m of the goal before 90~s. End-to-end success uses all 200 trials, conditional success uses only execution-eligible trials, and Native--RB CIs use 10,000 scene-stratified, query-clustered bootstrap resamples.

\subsection{Static Same-Graph Planning}

\begin{figure*}[!t]
    \centering
    \includeorplaceholder[width=0.95\textwidth]{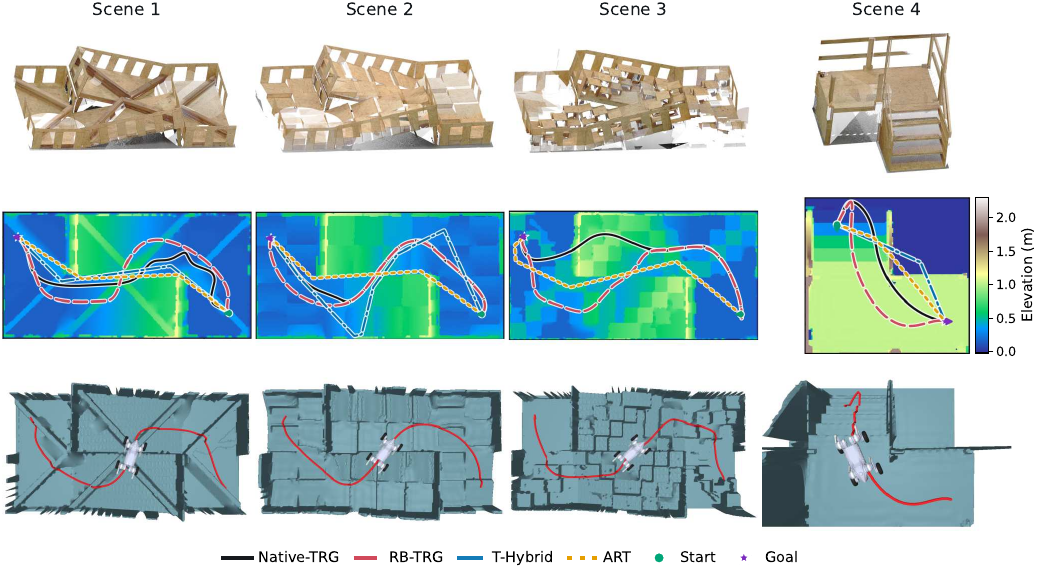}
    \caption{Qualitative results across Scenes~1--4. Rows show scanned point clouds, multi-method height-map routes for one representative query per scene, and representative closed-loop \method executions. T-Hybrid exhausts its 80,000-expansion budget without a route in Scene~3.}
    \label{fig:qualitative_results}
\end{figure*}

Relative to Native TRG, Full \method changes the node sequence in 157 of 200 cases and reduces averaged front--rear variation, lateral inclination, and interference by $10.0\%$, $8.5\%$, and $14.9\%$, respectively; untrusted exposure remains comparable. All four variants return routes for the same 200 cases, establishing matched paired coverage. At a $2.3\%$ path-length increase, Full \method records the lowest three core risks among all variants. Its improvements over Edge-only and Mean-only show that yaw-sweep modeling and upper-tail aggregation both contribute to route-level body-risk reduction. Figure~\ref{fig:qualitative_results} provides scene-level context for the resulting routes.

\subsection{End-to-End Benchmark}

Table~\ref{tab:simulation_results} reports task-level results, with missing routes and failed initialization counted as end-to-end failures.

\begin{table}[t]
\caption{Task-Level Benchmark and Closed-Loop Ablation}
\label{tab:simulation_results}
\centering
\scriptsize
\setlength{\tabcolsep}{2.0pt}
\resizebox{\columnwidth}{!}{%
\begin{tabular}{lccc}
\toprule
Method & Route cov. [\%] & Cond. succ. [\%] & E2E succ. [\%]\\
\midrule
\multicolumn{4}{l}{\textit{Task-level benchmark}}\\
Native TRG & \textbf{100} & 59.5 & 51.5\\
\method    & \textbf{100} & \textbf{76.1} & \textbf{68.5}\\
T-Hybrid   & 75.0 & 12.2 & 9.0\\
ART        & \textbf{100} & 5.0 & 5.0\\
\midrule
\multicolumn{4}{l}{\textit{RB-TRG component ablation}}\\
Edge-only  & \textbf{100} & 70.7 & \textbf{70.0}\\
Mean-only  & \textbf{100} & 72.5 & \textbf{70.0}\\
Full \method & \textbf{100} & \textbf{76.1} & 68.5\\
\bottomrule
\end{tabular}%
}
\end{table}

Native TRG and \method produce routes for all 40 queries. Across 200 paired executions, \method succeeds 137 times versus 103 for native TRG, increasing end-to-end success by 17.0 percentage points (95\% scene-stratified, query-clustered bootstrap CI: 8.0--26.0 points). On the 34 queries with execution-eligible trials for both, success is $78.8\%$ for \method and $60.6\%$ for native TRG; the scene-macro paired difference is 14.0 points (95\% CI: 4.0--24.0 points).

\method improves or matches Native TRG in every scene: end-to-end success rises from $76.0\%$ to $94.0\%$ in Scene~1, remains $100\%$ in Scene~2, rises from $18.0\%$ to $66.0\%$ in Scene~3, and increases from $12.0\%$ to $14.0\%$ in Scene~4. The largest gain occurs in Scene~3.

Failure decomposition contextualizes the reproduced baseline results. With the consistently applied $0.42$~rad threshold, T-Hybrid returns routes for 30 of 40 queries. All ten planning abstentions occur in Scene~3 and account for 50 nominal trials; among the remaining 150 trials, two fail initialization, 130 reach the 90-s timeout, and 18 succeed. ART returns a route for all 40 queries, but 190 of its 200 executions time out and 10 succeed. Among execution-eligible trials, mean non-wheel-contact time is $41.5\%$ for T-Hybrid and $27.3\%$ for ART, while the mean per-trial 95th-percentile absolute pitch is $32.7^\circ$ and $43.6^\circ$, respectively. Route coverage alone therefore does not explain the lower success rates; most failures occur during closed-loop execution. Both planner families were developed for unstructured outdoor or field rough-terrain navigation~\cite{liu2023hybrid,wellhausen2023artplanner}, whereas the rescue courses evaluated here concentrate steep slopes, steps, narrow elevated structures, and large attitude changes within compact scenes. These results should therefore be interpreted as a severe cross-domain stress test, rather than as a general ranking of the planners.

The lower block of Table~\ref{tab:simulation_results} reports the closed-loop ablation on the same 40 queries and five friction-only seeds under Native fixed-mode perturbations. Full \method attains the highest conditional success ($76.1\%$ versus $70.7\%$ and $72.5\%$), whereas Edge-only and Mean-only each reach $70.0\%$ end-to-end success, 1.5 points above Full. Because execution eligibility is assessed after route-tangent initialization, conditional and end-to-end rankings need not coincide. The paired end-to-end differences are not statistically resolved: Full-minus-Edge-only and Full-minus-Mean-only are both $-1.5$ points, with bootstrap 95\% CIs of $[-7.5,5.0]$ and $[-8.5,6.5]$, respectively. Together with Table~\ref{tab:static_results}, these results support complementary contributions from yaw-sweep modeling and upper-tail aggregation.

Table~\ref{tab:paired_metrics} summarizes 170 trials eligible for both planners. Here, $\Delta$ denotes \method{} minus native TRG and lower values are better. Contact is the trial-duration percentage with non-wheel terrain contact; wheel slip is the mean normalized mismatch between wheel-surface and base forward speed. Best values are bold.

\begin{table}[t]
\caption{Paired Closed-Loop Diagnostics}
\label{tab:paired_metrics}
\centering
\scriptsize
\setlength{\tabcolsep}{1.8pt}
\begin{tabular}{lrrrr}
\toprule
Metric & Native & \method & $\Delta$ & 95\% CI\\
\midrule
Cross-track RMSE [m]       & 0.383 & \textbf{0.074} & $-0.309$ & $[-0.744,-0.013]$\\
Non-wheel-contact time [\%] & 27.1  & \textbf{18.2}  & $-8.8$   & $[-13.5,-2.2]$\\
Wheel-slip proxy            & 0.645 & \textbf{0.607} & $-0.037$ & $[-0.056,-0.013]$\\
95th-percentile $|$pitch$|$ [$^\circ$] & 24.5 & \textbf{20.4} & $-4.0$ & $[-6.2,-0.8]$\\
95th-percentile $|$roll$|$ [$^\circ$]  & 9.7  & \textbf{8.5}  & $-1.3$ & $[-3.0,0.3]$\\
\bottomrule
\end{tabular}
\end{table}

Point estimates favor \method for every diagnostic, and four of five confidence intervals exclude zero: cross-track error, non-wheel-contact time, wheel slip, and pitch. Roll also favors \method but has a wider interval. Together with the end-to-end gain, these diagnostics associate body-aware costs with more stable execution.

\begin{figure*}[!t]
    \centering
    \subfloat[Planned route in RViz]{%
        \includeorplaceholder[width=0.21\textwidth]{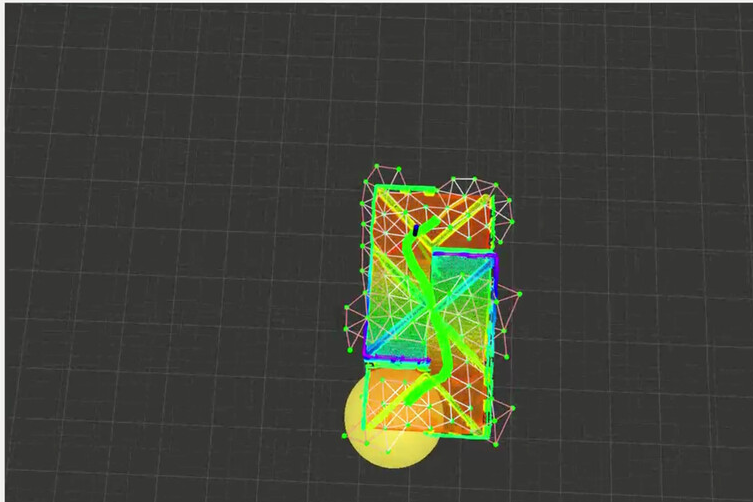}}\hfill
    \subfloat[LiDAR localization]{%
        \includeorplaceholder[width=0.21\textwidth]{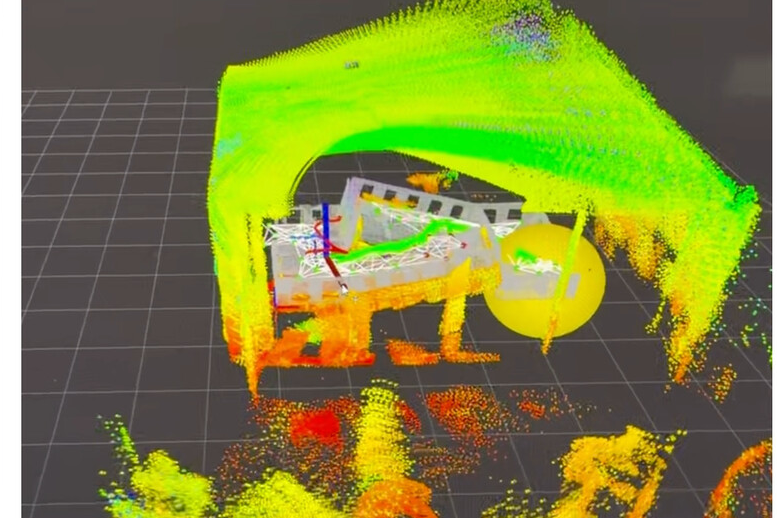}}\hfill
    \subfloat[Physical execution I]{%
        \includeorplaceholder[width=0.21\textwidth]{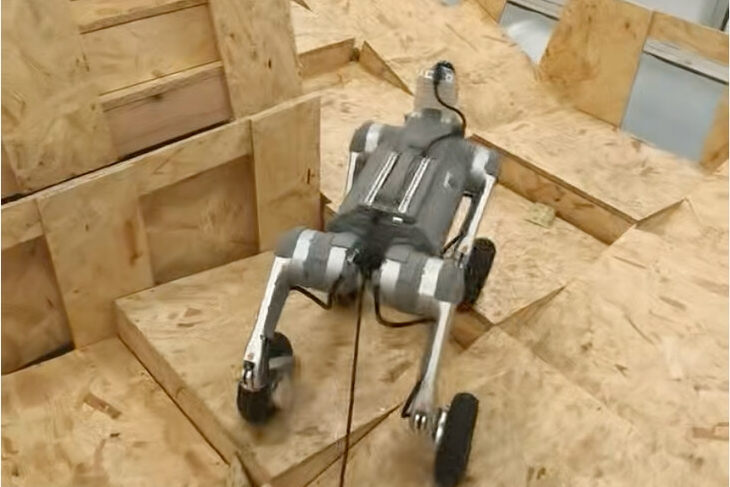}}\hfill
    \subfloat[Physical execution II]{%
        \includeorplaceholder[width=0.21\textwidth]{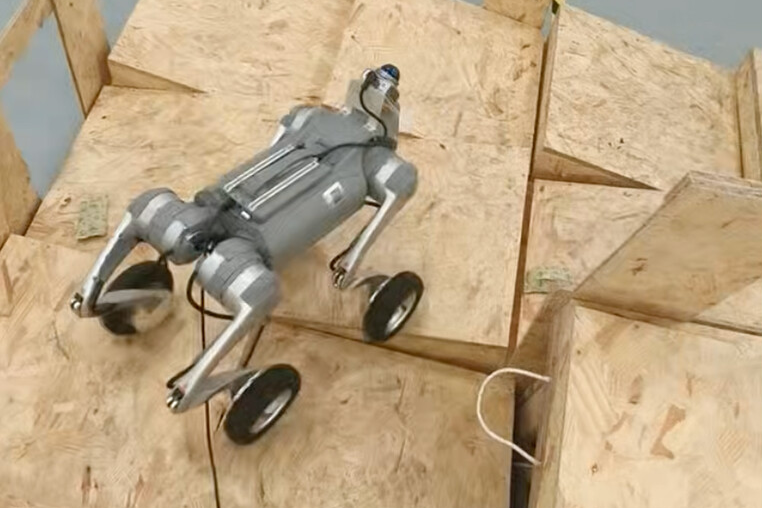}}
    \caption{Go2-W deployment. RViz views show the planned route and LiDAR localization; photographs show traversal of the stepped course.}
    \label{fig:real_world_deployment}
\end{figure*}

\subsection{Hardware Integration}

\method is integrated with the Unitree Go2-W LiDAR navigation stack shown in Fig.~\ref{fig:real_world_deployment}. The planned terrain-graph route and LiDAR localization are visualized in RViz, and the physical robot traverses the stepped-terrain course through the existing navigation and locomotion interfaces. The integrated system received the Best Mobility and Best Autonomy awards at the IEEE ICRA 2026 Legged Robot Challenges.

\section{Discussion}
\label{sec:discussion}

The static and closed-loop studies show that heading-conditioned body risk changes both route ranking and execution outcomes. On identical graphs, 157 route changes and the lowest averaged core body risks demonstrate the effect of the full formulation. In closed loop, the 17.0-point end-to-end gain and improvements in four of five paired diagnostics connect the selected routes to more stable execution.

The component ablations clarify how these gains arise. Yaw-sweep sampling captures body exposure during reorientation, while upper-tail aggregation preserves localized terrain hazards. Their combination yields the strongest static geometry and the highest conditional success, demonstrating distinct roles for the two mechanisms.

Because its transition costs are controller independent, \method serves as a modular route-selection component alongside graph construction, contact-feasibility analysis, and locomotion control. Its scope is route-level geometric risk rather than contact-level feasibility. Oriented footprint sampling increases per-query computation, while caching avoids repeated edge and turn evaluations. The controlled evidence covers four terrain scans, one Go2-W morphology, and a fixed locomotion policy; the hardware study demonstrates end-to-end system integration rather than a controlled performance comparison.

\section{Conclusion}
\label{sec:conclusion}

We presented \method, a heading- and turn-aware body-risk formulation for sparse TRG planning. Oriented footprints along graph edges and yaw sweeps change route ranking and reduce front--rear variation, lateral inclination, and terrain interference by $10.0\%$, $8.5\%$, and $14.9\%$, respectively, with a $2.3\%$ increase in path length. Against Native TRG, \method raises end-to-end success by 17.0 points and reduces tracking error, non-wheel-contact time, wheel slip, and pitch. Component ablations establish the distinct roles of yaw-sweep modeling and upper-tail aggregation, while the Go2-W deployment demonstrates integration with an existing navigation stack. Together, these results show that body-scale transition reasoning enables compact terrain graphs to select routes better aligned with closed-loop execution.

\bibliographystyle{IEEEtran}
\bibliography{references}

\end{document}